\documentclass[10pt,twocolumn,letterpaper]{article}

\usepackage[pagenumbers]{cvpr} 

\usepackage{graphicx}
\usepackage{amsmath}
\usepackage{amssymb}
\usepackage{booktabs}
\usepackage{xcolor}
\usepackage{comment}
\usepackage{scalerel}
\usepackage{enumitem}
\usepackage[accsupp]{axessibility} 

\usepackage[pagebackref,breaklinks,colorlinks]{hyperref}

\usepackage[capitalize]{cleveref}
\crefname{section}{Sec.}{Secs.}
\Crefname{section}{Section}{Sections}
\Crefname{table}{Table}{Tables}
\crefname{table}{Tab.}{Tabs.}

\newcommand{\thor}{\textsc{AI2Thor}\xspace}
\newcommand{\pnav}{\textsc{PointNav}\xspace}
\newcommand{\onav}{\textsc{ObjectNav}\xspace}
\newcommand{\framework}{i\textsc{See}\xspace}

\newcommand{\rnon}{$RN_{\scaleto{ON}{3pt}}$\xspace}
\newcommand{\rnonrnd}{$RN_{\scaleto{ON}{3pt}}^{\scaleto{r}{4pt}}$\xspace}
\newcommand{\scon}{$SC_{\scaleto{ON}{3pt}}$\xspace}
\newcommand{\sconrnd}{$SC_{\scaleto{ON}{3pt}}^{\scaleto{r}{4pt}}$\xspace}
\newcommand{\rnpn}{$RN_{\scaleto{PN}{3pt}}$\xspace}
\newcommand{\rnpnrnd}{$RN_{\scaleto{PN}{3pt}}^{\scaleto{r}{4pt}}$\xspace}
\newcommand{\scpn}{$SC_{\scaleto{PN}{3pt}}$\xspace}
\newcommand{\scpnrnd}{$SC_{\scaleto{PN}{3pt}}^{\scaleto{r}{4pt}}$\xspace}

\def\cvprPaperID{7404} 
\def\confName{CVPR}
\def\confYear{2022}

\begin{document}

\title{What do navigation agents learn about their environment?}

\author{Kshitij Dwivedi, Gemma Roig\\
Goethe University Frankfurt \\
{\tt\small dwivedi@em.uni-frankfurt.de, roig@cs.uni-frankfurt.de}
\and
Aniruddha Kembhavi, Roozbeh Mottaghi\\
PRIOR @ Allen Institute for AI\\
{\tt\small anik@allenai.org, roozbehm@allenai.org}
}
\maketitle

\begin{abstract}

Today's state of the art visual navigation agents typically consist of large deep learning models trained end to end. Such models offer little to no interpretability about the learned skills or the actions of the agent taken in response to its environment. While past works have explored interpreting deep learning models, little attention has been devoted to interpreting embodied AI systems, which often involve reasoning about the structure of the environment, target characteristics and the outcome of one's actions. In this paper, we introduce the Interpretability System for Embodied agEnts (\framework) for Point Goal and Object Goal navigation agents. We use \framework to probe the dynamic representations produced by these agents for the presence of information about the agent as well as the environment. We demonstrate interesting insights about navigation agents using \framework, including the ability to encode reachable locations (to avoid obstacles), visibility of the target, progress from the initial spawn location as well as the dramatic effect on the behaviors of agents when we mask out critical individual neurons. The code is available at:
\url{https://github.com/allenai/iSEE}

\end{abstract}

\section{Introduction}
\label{sec:intro}

The research area of Embodied AI -- teaching embodied agents to perceive, communicate, reason and act in their environment -- continues to receive a lot of interest from the computer vision, natural language processing and robotics communities. A growing body of work has resulted in the emergence of several powerful and visually rich simulators including AI2-THOR~\cite{ai2thor}, Habitat~\cite{habitat19iccv} and iGibson~\cite{Shen2020iGibsonAS}; works that require agents to navigate~\cite{Anderson2018OnEO}, reason~\cite{rearr}, collaborate~\cite{jain2020cordial}, manipulate~\cite{Ehsani2021ManipulaTHORAF} and follow instructions~\cite{anderson2018vision}.

\begin{figure}[t]
  \centering
   \includegraphics[width=\linewidth]{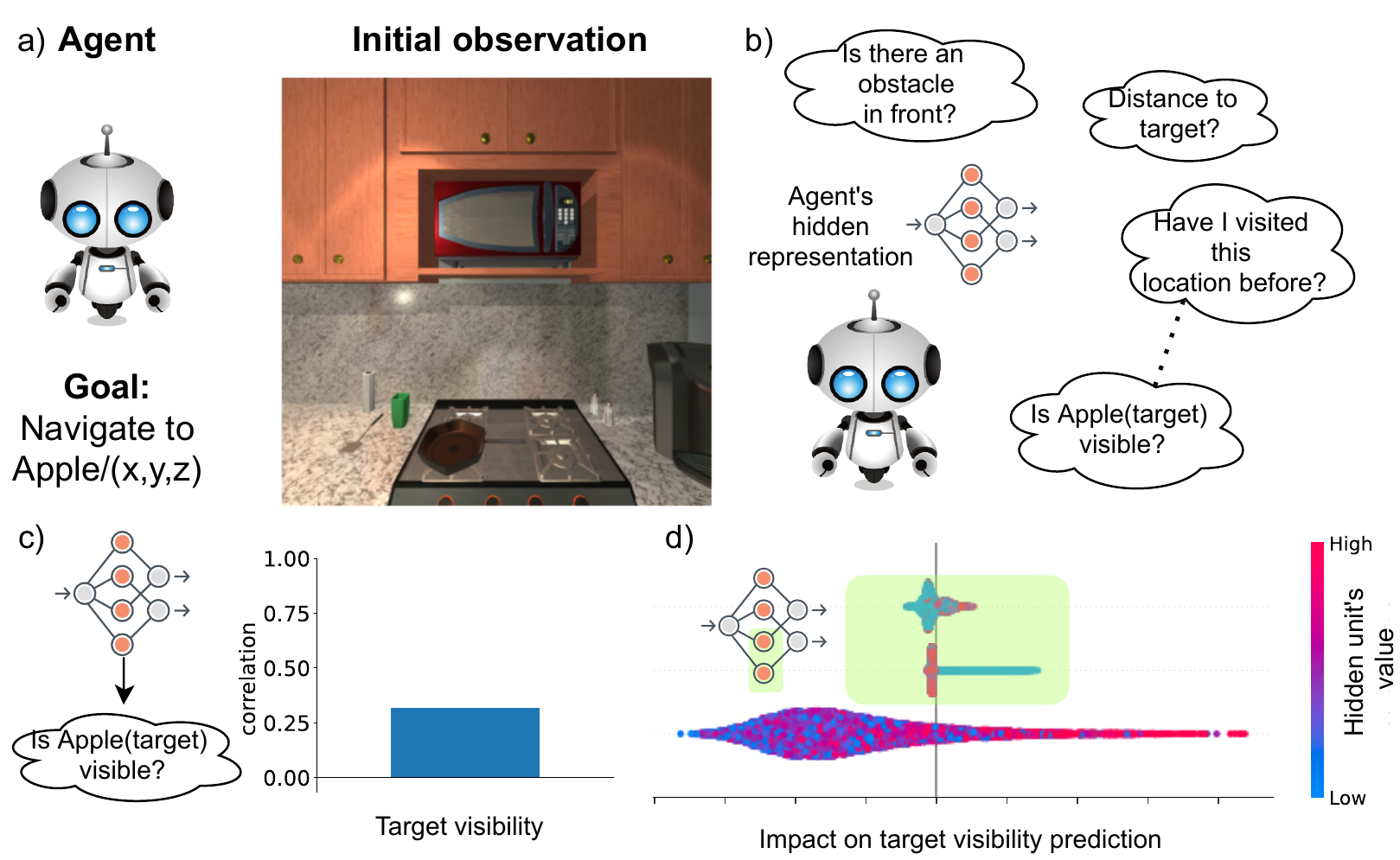}
   \vspace{-0.5cm}
   \caption{\textbf{The \framework framework.} (a) An agent learns to perform the \onav\ or \pnav\ tasks. (b) We wish to explore what information is encoded in the hidden representations of the agent. (c) To achieve this, we evaluate how well the agent's hidden representation can predict human interpretable concepts e.g. target visibility in ObjectNav. (d) Then we apply an explainablity method SHAP~\cite{lundberg2020local2global} to identify the top-k relevant units.}
   \vspace{-0.8cm}
   \label{fig1}
\end{figure}

While fast progress is being made across a variety of tasks and benchmarks, most solutions being employed are black box neural networks trained to either imitate a sequence of human/oracle actions or trained via reinforcement learning with a careful selection of positive and negative rewards. These models offer little to no interpretability out-of-the-box about the concepts and skills learned by the model or about the actions taken by the model in response to a task or observation. Developing interpretable systems is particularly important in embodied AI since we expect these systems to eventually be deployed onto robots that will navigate the real physical world and interact with people in it. 

In the image classification literature, a number of interpretability methods have been developed over the past few years~\cite{yosinski2015understanding,bau2017network,nguyen2016synthesizing,fong2018net2vec}. These methods rely on probing model activations via various inputs or generating synthetic inputs that lead to a spike in an activation. While such methods are useful in probing Embodied AI models, they do not take into account the rich metadata (such as perfect segmentation, depth maps, precise object localization, etc.) available in synthetic environments commonly used to train these models. Simulated worlds provide us a unique opportunity to expand interpretability research to embodied agents and develop new methods that take advantages of rich metadata.

We propose a framework to interpret the hidden representations of embodied agents trained in simulated worlds. We apply our framework to two navigation tasks (Figure \ref{fig1}a): Object Navigation (\onav)~\cite{Batra2020ObjectNavRO}, the task of navigating to a target object and Point Goal Navigation (\pnav)~\cite{Anderson2018OnEO}, the task of navigating to a specified relative co-ordinate, within the \thor\ environment; but our methods are general and can be easily applied to more tasks and other environments. We train agents to perform these tasks and then probe their hidden representations to evaluate if they encode aspects of their task, progress and surroundings (Figure \ref{fig1}b and \ref{fig1}c). We then apply the model interpretation method SHAP~\cite{lundberg2020local2global} to identify which hidden units are most relevant for predicting these concepts (Figure \ref{fig1}d). Our framework allows us to gather evidence towards answering two fundamental questions about a trained model: (1) Has the model learned a particular concept ? (2) Which units within a recurrent layer encode this concept ? Using this framework, we were able to find several interesting insights about \onav\ and \pnav agents.

The key contributions of this work are:
\begin{itemize}[noitemsep,topsep=0pt,parsep=0pt,partopsep=0pt]
  \item A new interpretability framework specialized for navigation agents with no linearity assumptions between concepts and hidden units.
  \item New insights about what navigation agents encode and in which units:
  \begin{itemize}[noitemsep,topsep=0pt,parsep=0pt,partopsep=0pt]
      \item sparse target representation in \onav (50/512 units) and \pnav (5/512 units);
      \item learning of concepts such as reachable locations and visit history by \onav agents; encoding of progress towards target and less reliance on visual information by \pnav agents.
  \end{itemize} 
  \item Ablation experiments showing no impact on model performance after removal of 10\% units suggesting redundancy in the representation.
\end{itemize}

\section{Related Work}
\label{sec:related}
We explore representations stored within an agent's hidden units by predicting a human interpretable piece of information about the agent and its environment. Our work is related to two directions in interpretability research: (1) Interpretability of individual hidden units and (2) Explaining model’s predictions. 

\noindent \textbf{Interpretability of hidden units.} A common approach to investigate what a hidden unit encodes is to find the input image also referred to as ``preferred image" that leads to a maximal activation of the unit of interest. The preferred image can be from within the examples in a dataset~\cite{yosinski2015understanding,zeiler2014visualizing} or obtained using gradient descent by optimizing over the input~\cite{nguyen2016synthesizing,olah2017feature,wei2015understanding,erhan2009visualizing,simonyan2013deep,goh2021multimodal}. One disadvantage of the methods using preferred images is that it is difficult to quantify the association of a unit with a concept. To address this issue, NetDissect~\cite{bau2017network,zhou2014object} uses overlap of a unit's spatial activation with groundtruth segmentation maps of a human interpretable concept as a measure to quantify a unit's association with a concept. The idea was further extended in Net2vec~\cite{fong2018net2vec} to investigate whether a single unit or a group of units encode a concept. However, these approaches require groundtruth pixel-level annotation for every concept of interest and therefore for new concepts, new annotations are required. On the other hand, simulation environments \cite{ai2thor,habitat19iccv,Shen2020iGibsonAS} have annotations readily available as a part of the metadata. However, given the vast amount of metadata beyond simply object information, there is a need to develop new methods for these environments to interpret embodied agents. Recent embodied AI works\cite{Weihs2021LearningGV,ye2021auxiliary} have started focusing in interpretability by linear decoding of concepts from hidden units\cite{Weihs2021LearningGV} and finding computational structure of the agent's recurrent units using fixed point analysis\cite{ye2021auxiliary}. Patel et al. \cite{patel2021interpretation} explored interpretation of emergent communication in collaborative embodied agents. However these works do not focus on identifying which hidden units encode a given concept which is one of the main contributions of the present work.

\noindent \textbf{Explaining model predictions.}
Saliency methods\cite{selvaraju2017grad,smilkov2017smoothgrad,omeiza2019smooth,rebuffi2020saliency,bach2015pixel} use gradients to find which pixels of an image are relevant for model's prediction. Additive feature attribution methods \cite{ribeiro2016should,NIPS2017_7062,shrikumar2017learning} investigate the effect of adding an input feature in model prediction. A disadvantage of these methods is that they focus on explaining the model predictions on raw pixel level. To explain the model prediction using human-interpretable concepts, TCAV~\cite{kim2018interpretability} and subsequent works~\cite{ghorbani2019towards,goyal2019explaining,yeh2019completeness} were proposed that use concept vectors instead of raw pixels to explain model prediction. To find concept vectors additional human annotations are required. In the embodied environments~\cite{ai2thor,habitat19iccv,Shen2020iGibsonAS}, we have the advantage of already annotated human interpretable concepts.

The above two directions of research  have been considered as independent directions of interpretability research -- one focusing on interpreting what the hidden units learn and the other on interpreting the decisions made by the model. In this work, we observe the potential of linking two approaches to interpret what the hidden units learn by using human interpretable concepts. Specifically, we train an interpretable model (Gradient boosted Tree) to predict human intepretable concepts from the hidden units of the model and then apply a global model explainability method SHAP~\cite{lundberg2020local2global} to explain which units are relevant for which concept prediction. In this work, we use SHAP because (a) it provides a
unique solution with three desirable properties: local accuracy, missingness and consistency~\cite{NIPS2017_7062}, (b) it unifies several model agnostic~\cite{ribeiro2016should,shrikumar2017learning} and tree based explanation methods~\cite{treeint}, and (c) it provides explanation on both local (single example) and global (dataset) levels.

\noindent \textbf{Embodied tasks.} Several approaches have been proposed \cite{zhu2017target, wortsman2019learning,yang2018visual, Wijmans2020DDPPOLN,ramakrishnan2020occupancy,chaplot2020object, Shen2020iGibsonAS,wani2020multion,mousavian2019visual,Li_2020_CVPR,Chattopadhyay2021RobustNavTB,Mayo2021VisualNW,Du2020LearningOR,chen2021semantic} to tackle the navigation problem, which is a core task in Embodied AI. In this paper, we analyze standard base models for two popular navigation tasks, PointNav~\cite{Anderson2018OnEO} and ObjectNav~\cite{Batra2020ObjectNavRO}.

\section{Interpretability Framework}
\label{sec:method_description}
\begin{figure*}
  \centering
  \includegraphics[width=0.98\linewidth]{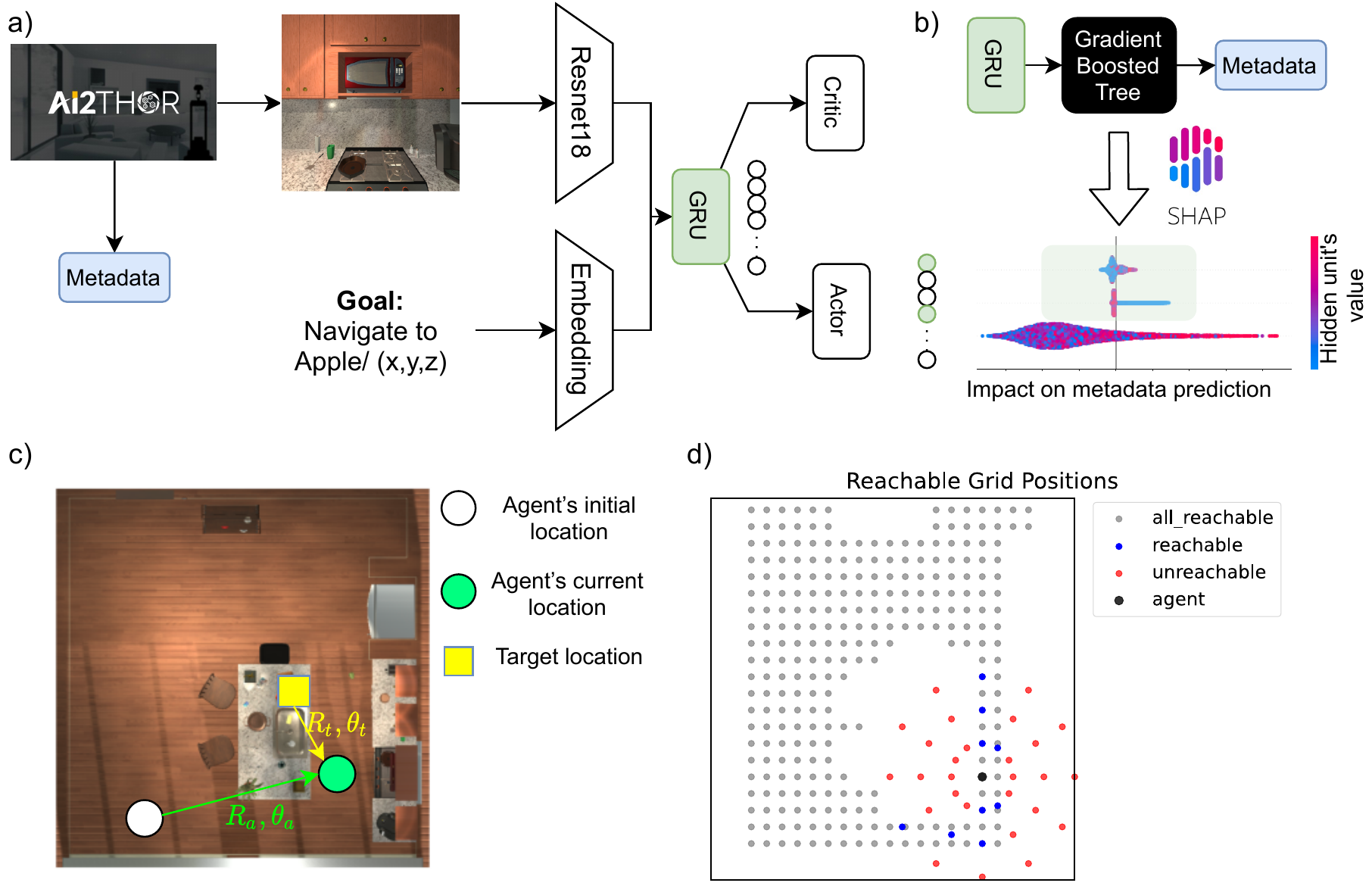}
  \vspace{-0.3cm}
  \caption{\textbf{\framework:} a) At a given timestep, \thor\ generates an observation that is fed as input to the agent along with a goal embedding. For that time step, we also extract relevant event metadata from \thor\ which is unseen by the agent. b) After sampling rollouts from multiple training and validation episodes, we train a gradient boosted tree to predict metadata from the agent's hidden representation (GRU units). We then apply SHAP, an explainability method that identifies the top-k most relevant units for predicting a given metadata type. c) At a given timestep, we extract agent's orientation with respect to its initial spawn location ($R_{a},\theta_{a}$) and target location ($R_{t},\theta_{t}$). d) We extract reachable positions at distance 2,4,6 times the grid size and different angles with step size of 30 degrees to identify whether these locations can be reached by the agent or not.}
    \vspace{-0.4cm}
  \label{fig2}
\end{figure*}

We introduce the \underline{\textbf{I}}nterpretability \underline{\textbf{S}}ystem for \underline{\textbf{E}}mbodied ag\underline{\textbf{E}}nts (\framework). \framework\ probes agents at their understanding of the task given to them, their progress at this task and the environment they act in. This probing is done via training simple machine learning models that input network activations and output the desired information. Simulated environments provide us with a gamut of metadata about the agent, task and surroundings, allowing us to train a series of models for probing this information. \framework\ also helps identifying specific neural units that store this information. This is done via computing the SHapley Additive exPlanations (SHAP)~\cite{shapley201617} values for individual neural units. Finally we study the effect of switching off individual neural units on the downstream tasks that the agents are trained for.

We study embodied agents trained for \pnav~\cite{Anderson2018OnEO} (navigation towards a specific coordinate in a room) and \onav~\cite{Batra2020ObjectNavRO} (navigation towards a specific object). Our agents encode their visual observations via a convolutional neural network and encode their target/goal via an embedding layer. The outputs of the visual and goal encoders are fed into a gated recurrent unit (GRU) to add memory. The hidden units of the GRU are then linearly transformed into the policy (distribution over actions) (Figure \ref{fig2}a). There are more complex, customized models for each of these tasks that achieve higher performance. However, we utilize these simple, generic models that can be applied to various tasks and make the comparisons across tasks more fair. In this work, we use \framework\ to probe the hidden units in the GRU and use gradient boosted trees (GBT) as the ML model to determine the presence of relevant information within these hidden units (Figure \ref{fig2}b). We focus here specifically on GRU units since (a) we are interested in analyzing dynamic visual representations (GRU units) as opposed to static visual representations (CNN visual encoder) and (b) some of our models use a frozen visual encoder and only optimize the parameters within the GRU.

We now describe the metadata extracted from the simulator, probing for this metadata via building GBTs and using SHAP to identify individual hidden units that store the relevant information.

\subsection{Metadata}
We probe agents at their understanding of the target, their position in the scene, the reachability of objects in their surroundings and their memory of visited locations as they navigate their world. This information is easily extracted by us from the metadata provided by the simulator.

\textbf{Target Information:} Agents trained for the \onav\ and \pnav\ tasks must navigate to the location of a specified object or a point, respectively. In either case, one might expect an agent to be able to estimate its positioning with respect to the goal. Therefore, at a given timestep $t$, we extract metadata containing the distance ($R_{t}$) and orientation ($\theta_{t}$) of the agent from the target (Figure \ref{fig2}c). In \onav, an agent is successful if the object lies within 1m of the agent and is visible; thus we additionally extract target visibility ($visible_{t}$). Since an object may be visible in the frame but not within the specified distance to determine success, we also extract the percent of pixels covered by the target object using segmentation masks provided by AI2-THOR ($Area_{t}$). 

\textbf{Agent's information:} Memory of how far and in what direction one has travelled can be relevant to avoiding re-visiting locations in the scene. Therefore, we extract the agent's distance ($R_{a}$) and orientation ($\theta_{a}$) with respect to its starting location (Figure \ref{fig2}c). 

\textbf{Reachability:} For an agent to successfully navigate in a scene it should be able to detect obstacles and its path around them. Thus, we extract metadata to detect whether a particular location with respect to the agent's current location is reachable or not. Given an agent's location, we first extract all reachable gridpoints in the scene. Then, with the agent's location as the center we consider three concentric circles with radii=2, 4,and 6 times the grid size and locate points on these circle that are at angles from 0 to 360 in the steps of the agents rotation angle (=30 degrees). For each of these points $R_{r},\theta_{angle}$, where $r$ is the radius and $angle$ is the orientation of the grid point with respect to agent in degrees, we check whether the closest reachable gridpoint is within $gridSize/\sqrt{2}$ or not. Figure \ref{fig2}d illustrates such reachable gridpoints in the scene.

\textbf{Visited History:} The metadata extracted above captures a global summary of the agent's movements. We also extract its local visit history. This is done by checking if a location  ($\texttt{visited}\,_{l}$), rotation ($\texttt{visited}\,_{lr}$) and camera horizon ($\texttt{visited}\,_{lrh}$)  has been visited by the agent or not. 

\subsection{Metadata extraction}
As the agent traverses around in a scene, we extract the GRU activations of the agent along with the agent and scene metadata described above. This is done within the training and validation scenes. The latest model architectures and training algorithms for \pnav\ and \onav\ lead to very capable agents that (a) exhibit little variability in their trajectories (b) do not collide often (c) make few mistakes such as revisiting locations. Such trajectories are less useful to probe agents, since the events of interest occur sparsely. Hence we use human trajectories (trajectories specified by humans navigating around) that encourage exploration and have intentional collisions and mistakes. Using a pre-defined set of human trajectories also enables us to fairly compare findings across agents.

\subsection{Metadata prediction}
We train GBTs to predict specific metadata concepts using the GRU's hidden units as inputs. GBTs are trained using episodes within the training scenes and evaluated using correlation between the predicted metadata and groundtruth metadata on the validation episodes. For a given model, we trained one GBT of $depth=10$ for each concept using xgboost library. For binary variables (such as target visibility) we use the logistic loss function and for continuous variables (such as distance from target/agent's initial position) we use the mean squared error loss function. Total training and evaluation time of GBT was 8 seconds on a single NVIDIA RTX 2070 GPU. We use GBTs because: (1) they are more interpretable in comparison to many other ML models when the mapping from inputs to outputs is not linear; (2) allow exact computation of SHAP values as compared to other models where SHAP values can only be approximated \cite{lundberg2020local2global}.

\begin{figure}
  \centering
  \includegraphics[width=1.0\linewidth]{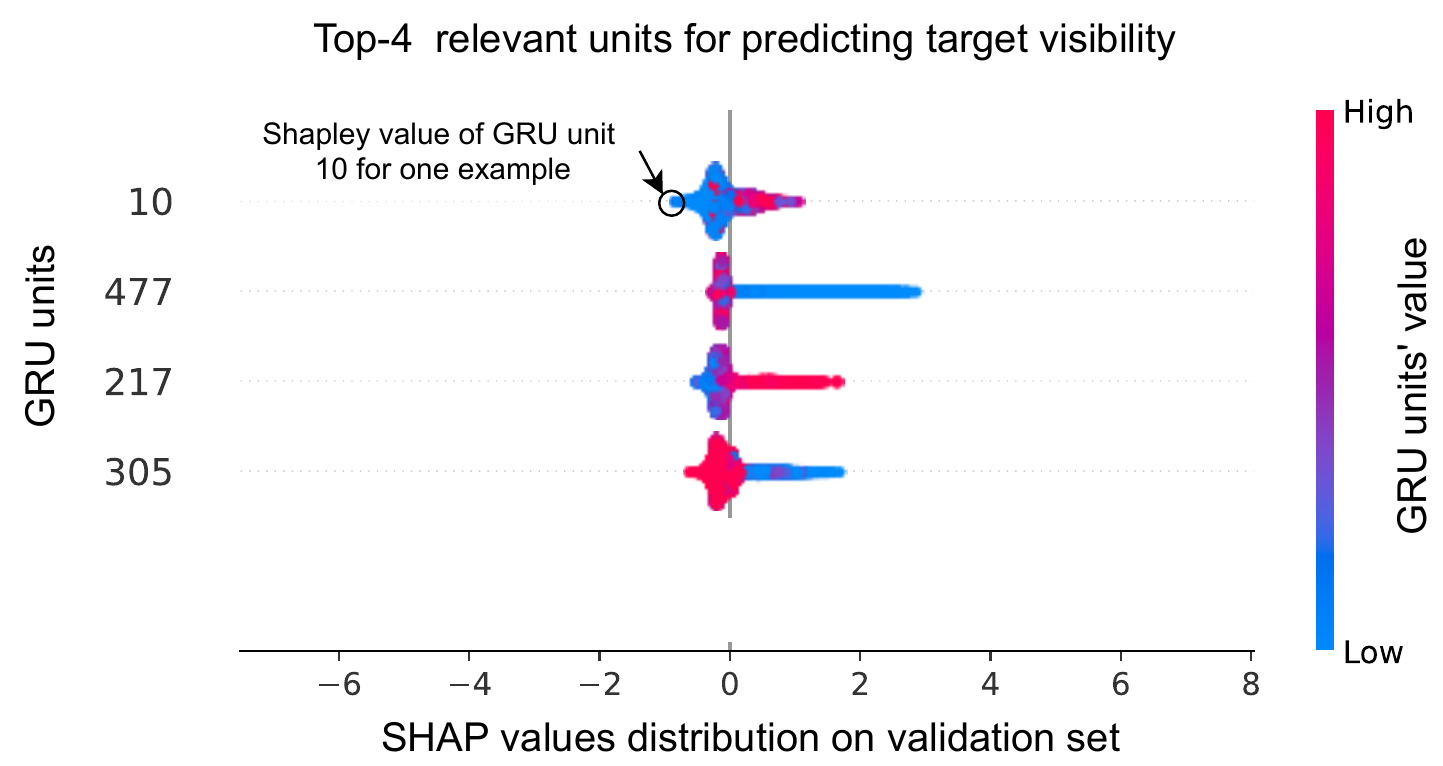}
  \vspace{-0.5cm}
  \caption{\textbf{Schematic to read a SHAP plot:} The plot shows the top-4 relevant GRU units to predict the target visibility. Each row shows the distribution of SHAP values of a given GRU unit for all the examples in the validation set with each dot in the row corresponding to an individual data point. The color of the dot indicates whether the GRU unit's output was low or high for that data point.}
  \vspace{-0.5cm}
  \label{fig3}
\end{figure}

\subsection{Identifying explainable units using SHAP}
Given a set of hidden units, SHAP computes the importance of each individual unit by quantifying its contribution towards predicting a concept. SHAP values are based on a game theory concept called Shapley values \cite{shapley201617}. We first train a GBT to predict a concept using all hidden units. We then use a subset of hidden units and mask other units to predict a concept using pretrained GBT. Then we add in a new hidden unit and compute the change in the model's prediction capability. This difference quantifies the contribution of a hidden unit with regards to the chosen subset. By averaging this contribution over all possible subsets of hidden units, we get the Shapley value of the unit of interest. For instance, we use this method to compute the contribution of a specific GRU hidden unit towards predicting the visibility of the specified target. Note that the obtained Shapley value indicates the impact of the hidden unit on the model's outcome for a single example. To quantify the global impact of hidden unit on model's outcome we calculate the mean of absolute SHAP values over all examples in the validation set (for more
details please see Appendix \ref{A1}).  

Figure~\ref{fig3} is a SHAP beeswarm plot to visualize the global contribution of the top-k relevant GRU units. We use this plot to explain how one can interpret SHAP plots. This plot visualizes the contribution of the top 4 relevant units to predict target visibility. Each row corresponds to a given GRU unit, and each dot in the row corresponds to the GRU unit's Shapley value for a given example. Each row displays the distribution of SHAP values on all the samples of the validation set. The location of a dot on the x axis shows whether the impact of the GRU unit on model's prediction (i.e. Shapley value) is positive or negative. The GRU unit's value for a sample is visualized using the colorbar on right. As an example, for the circled dot in Figure~\ref{fig3}, the Shapley value of GRU unit 10 is negative and the color of the dot indicates that GRU unit 10's value is also low. For the examples on the right side of x-axis the shapley values are positive and the GRU unit's values are also higher. This means that GRU unit 10 is positively correlated with target visibility. Using a similar logic GRU unit 477 seems to be negatively correlated with target visibility. In a nutshell, the SHAP plot shows the global contribution of a GRU unit in prediction of a concept (rows sorted by contribution), displays the distribution over the validation examples (points in each row) and indicates whether a unit is positively or negatively correlated with the concept (colors of the points in accordance with the x-axis values).

\section{Experimental Setup}

\begin{figure}
  \centering
  \includegraphics[width=1.0\linewidth]{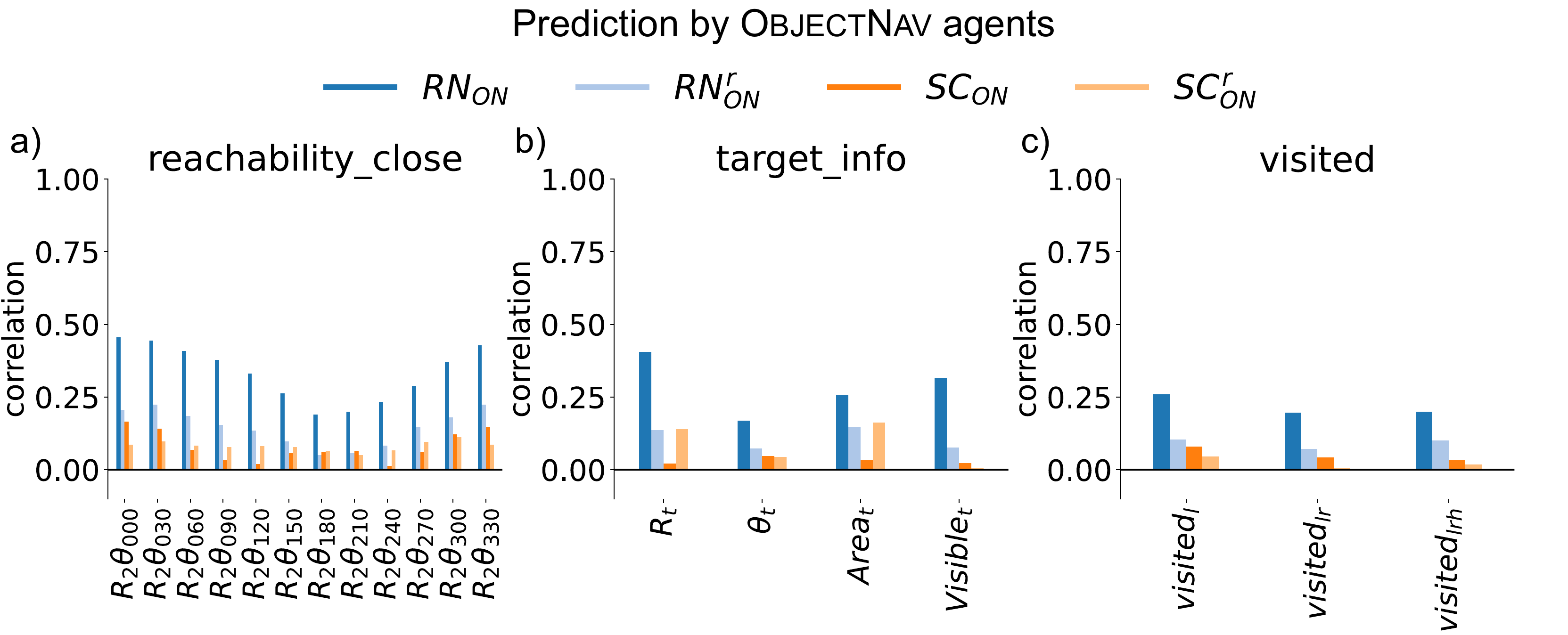}
  \vspace{-0.5cm}
  \caption{\textbf{Metadata prediction by \onav GRU units:} a) Reachability b) Target information c) Visited history}
  \label{fig4}
\end{figure}
\label{sec:experiments}
We use the AllenAct~\cite{AllenAct} framework to train models for the tasks \onav\ and \pnav\ tasks in the iTHOR rooms within \thor~\cite{ai2thor}. For both tasks, we use the same split of rooms for training and validation.

\subsection{\onav\ Models and Baselines}
We consider two models for \onav. The first model uses a frozen ResNet18 as the visual encoder and is named \rnon, while the second uses a 5 layer CNN (referred to as SimpleConv) as the visual encoder, denoted by \scon. In \scon, the visual encoder is optimized using the gradients of the actor critic loss. The visual representation is concatenated with the goal embedding which is then fed to a GRU. The GRU is connected to two linear layers predicting the policy and value. To ascertain if the representations learned by \onav\ agents are due to training, we consider two randomly initialized models with the same architetcures as the baselines. For the random ResNet model, named \rnonrnd, we initialize ResNet with ImageNet weights and initialize the GRU randomly. For the random SimpleConv model, named \sconrnd, both the visual encoder and GRU are initialized randomly. \rnon\ and \scon\ are trained for 300 Million steps using the default hyperparameters from the AllenAct framework.

\subsection{\pnav\ Models and Baselines}
Similar to \onav\ models we consider a ResNet based model (\rnpn) and a SimpleConv based model (\scpn). The distance and orientation to target are used as a sensory input to the model for target information. The corresponding random baselines are named \rnpnrnd\ and \scpnrnd. \rnpn\ and \scpn\ are trained for 300 Million steps using the default hyperparameters from AllenAct. 
\
\begin{center}
\footnotesize
\begin{tabular}{ ccccc } 
 \hline
 & \multicolumn{2}{c}{\textbf{ObjectNav}} & \multicolumn{2}{c}{\textbf{PointNav}}\\
 \hline
  & ResNet18 & SimpleConv & ResNet18 & SimpleConv \\
 \hline
 Trained & \rnon & \scon & \rnpn & \scpn \\
 Random & \rnonrnd & \sconrnd & \rnpnrnd & \rnpnrnd \\ 
 \hline
\end{tabular}
\end{center}

\begin{figure*}
  \centering
  \includegraphics[width=.99\linewidth]{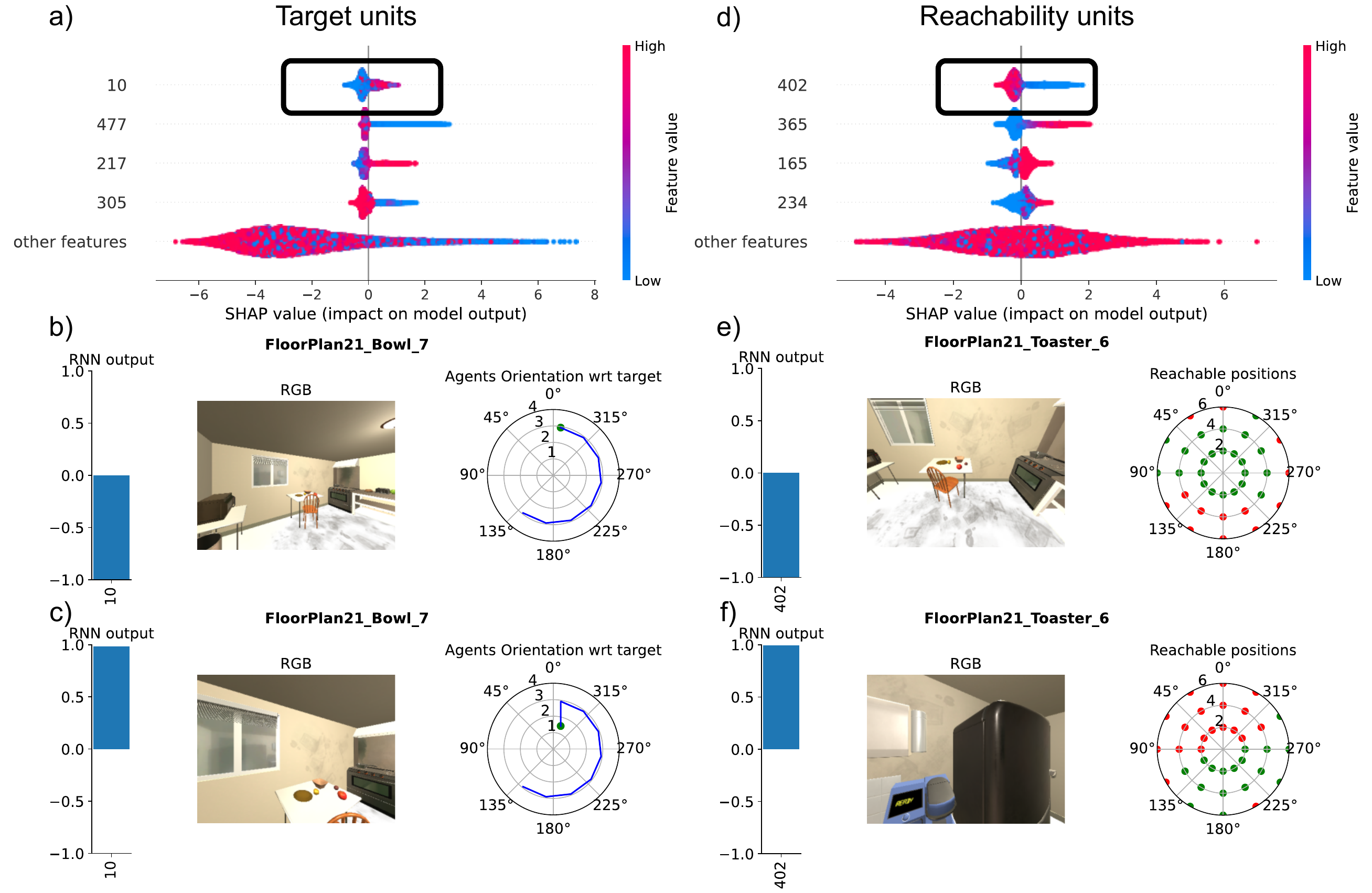}
  \vspace{-0.3cm}
  \caption{\textbf{Visualization of hidden units.} a) Target Visibility Unit:  Top-4 most relevant units to predict target distance. b) The bar plot on left shows unit 10's (target unit) response. The center image is agent's current observation. The polar plot on right shows the distance (in meters) and orientation of the agent (in degrees) wrt target. In this case, the agent is at around 3 meters away from the target and is oriented around 0 degrees. The response of unit 10 is negative. c) In this case, the agent is now closer to target (around 1 meters) and unit 10's (target unit) response is positive. d) Reachability Unit: Top-4 most relevant hidden units to predict reachability at distance $2\times gridSize$ and theta zero. e) The bar plot on the left shows unit 402's (reachability unit) response. The polar plot on right shows if the locations at radii of $2,4,6 \times gridSize$ and a given orientation in degrees are reachable or not. In this case, all the locations ahead are green i.e. reachable. The response of unit 402 is negative. f) All the locations ahead are red i.e. not reachable. The response of unit 402 is positive.}
  \vspace{-0.3cm}
  \label{fig5}
\end{figure*}

\subsection{Human Trajectories}
After training the \onav\ and \pnav\ models, 
we collect human sampled trajectories for the training and validation rooms. The training trajectories contain 59 episodes with average episode length of 480 while validation trajectories contain 42 episodes with average episode length of 470. The subject was encouraged to completely explore the rooms with intentional collisions and visits to previously visited locations with an episode length upper limit  of 500. All 8 models are forced to follow these trajectories. The corresponding metadata and GRU activity was extracted resulting in 28,000 training samples and 20,000 validation samples for GBT training.
\vspace{-0.2cm}
\section{Results}
\label{sec:results}
\subsection{\onav}
\vspace{-0.2cm}
The validation performance of \onav\ models saturates at around 50 million steps, therefore we select a checkpoint right after 50 million steps from both models. \rnon\ (success = 0.458, SPL = 0.23) significantly outperforms \scon\ (success = 0.124, SPL = 0.056). Here success indicates the fraction of episodes the agent successfully reached the target and SPL refers to Success
weighted by Path Length introduced in \cite{Anderson2018OnEO}. We consider concepts derived from metadata that are related to target information ($R_{t}$,$\theta_{t}$,$\texttt{visible}\,_{t}$,$\texttt{Area}\,_{t}$) , reachability ($R_{r}\theta_{angle}$ where $r$ is the radius and angle is the orientation of the neighboring grid point w.r.t. the agent), agent's information ($R_{a}$,$\theta_{a}$) and visited history ($\texttt{visited}\,_{l}$,$\texttt{visited}\,_{lr}$,$\texttt{visited}\,_{lrh}$).

\textbf{Metadata prediction:} We train GBTs to predict metadata from the GRU units. We observe that \rnon\ predicts reachability much better than the other three \onav\ models (Figure \ref{fig4}a) with a correlation of 0.45 and ROC\_AUC=0.75 for reachability in front ($R_{2}\theta_{000}$). We also observe an interesting pattern that prediction of reachability drops as one moves from 0 (front of the agent) to 180 (behind) degree then it starts increasing from 180 to 330 degrees suggesting the reachability of locations in front is more predictable than behind the agent. In Figure \ref{fig4}a, we show the results for reachability with radius = $2\times gridsize$. We observe a similar pattern for radius = $4\times gridsize$ and radius = $6\times gridsize$ (refer to Appendix \ref{A2}).

For the target information ($R_{t}$,$\theta_{t}$,$\texttt{Area}\,_{t}$,$\texttt{visible}\,_{t}$) \rnon\ shows a higher correlation than the other three models (Figure \ref{fig4}b). Visited history also ($\texttt{visited}\,_{l}$,$\texttt{visited}\,_{lr}$,$\texttt{visited}\,_{lrh}$) shows a higher correlation for \rnon\ (Figure \ref{fig4}c). The agent's information ($R_{a}$,$\theta_{a}$) is not predicted well and \rnon\ model shows a correlation similar to baselines suggesting this information is not learned by the agent during training (refer to Appendix \ref{A2}). Overall, we observe that \rnon\ learns the reachability, target relevant information and visited history from \onav\ training. This suggests that these three features are very crucial for performing this task.

While we present the results only on four concepts we also considered collision but found that it was poorly predicted for all 4 models (refer to Appendix \ref{A2}). 
\begin{figure}[tp]
  \centering
  \includegraphics[width=1\linewidth]{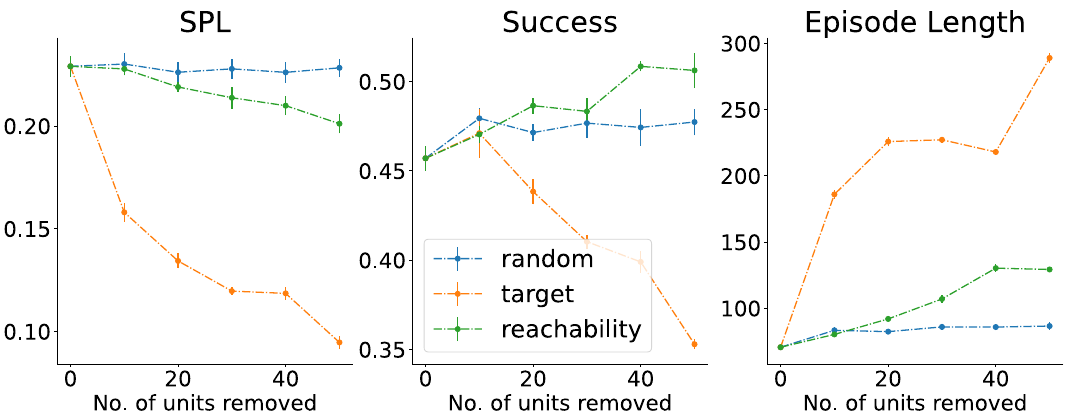}
  \vspace{-0.6cm}
  \caption{\textbf{Impact of removing units from \rnon.}}
  \vspace{-0.3cm}
  \label{fig6}
\end{figure}

\begin{figure*}[tp]
  \centering
  \vspace{-0.3cm}
  \includegraphics[width=1\linewidth]{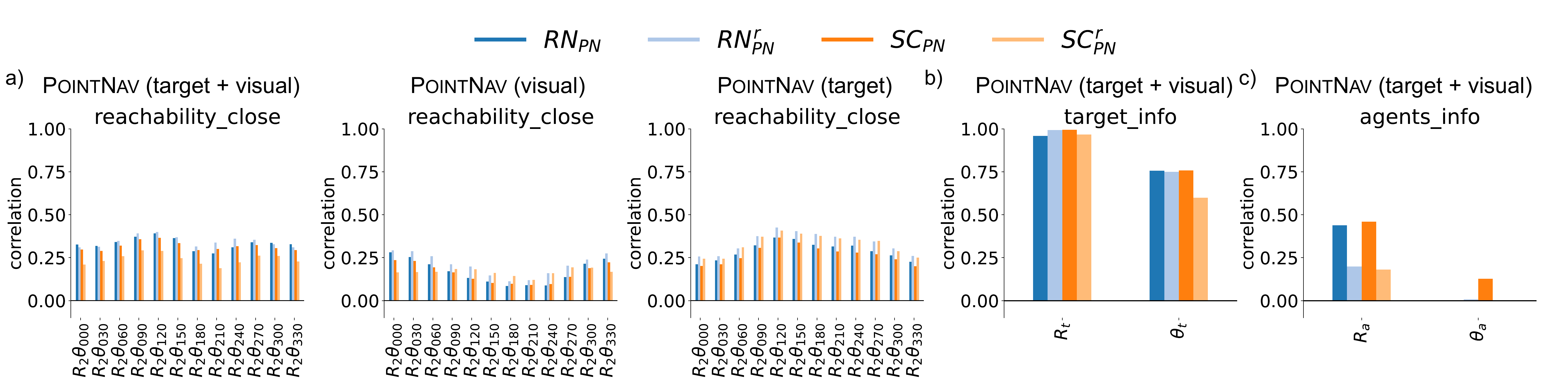}
  \vspace{-0.5cm}
  \caption{\textbf{Metadata prediction by \pnav GRU units:} a) Reachability b) Target information and c) Agent information }
  \vspace{-0.3cm}
  \label{fig7}
\end{figure*}

\textbf{Hidden unit visualization:}
To identify which hidden units are relevant to the mentioned concepts we apply SHAP on the two most interesting concepts ($\texttt{visible}\,_{t}$ and $R_{2}\theta_{000}$). In Figure~\ref{fig5}a we show the top-4 units that are most relevant in predicting the target visibility. On observing the SHAP plot of unit 10 (Figure \ref{fig5}a) we see that when the unit's value is higher it has a positive impact on target visibility and vice-versa suggesting that the unit's value is high when the target is visible (for aggregate SHAP values over units see Appendix \ref{A5}). The polar plots show the agent's trajectory (Figure~\ref{fig5} b,c), blue line represents trajectory and green dot indicates the agent's current location wrt target. Bar plot shows the RNN unit's response for current observation. Here, the target is a bowl; when the agent is away from the target its response is negative (Figure \ref{fig5}b) and when it is closer its response is positive (Figure \ref{fig5}c). These results also suggest that this unit might be positively correlated to target visibility.

In Figure \ref{fig5}d, we show the top-4 units most relevant in predicting $R_{2}\theta_{000}$ (for distribution of aggregate SHAP values over units see Appendix \ref{A5}). On observing the SHAP plot of unit 402 (Figure \ref{fig5}d) we can see that when the unit's value is higher it has negative impact on $R_{2}\theta_{000}$ and vice-versa suggesting that the unit value is high when the location ahead is not reachable. In Figure ~\ref{fig5}e,f, the dots are located at $radii=2,4,6 \times stepsize$ from the agent and at angles from 0 to 330 in steps of 30$^{\circ}$, where 0 is the front of the agent. Dot color indicates if the location is reachable (green) or not (red). Here, when the location in front of the agent is reachable the unit's response is negative (Figure \ref{fig5}e) and when there is an obstacle in front the unit's response is positive (Figure \ref{fig5}f). These results suggest that this unit might be detecting obstacles ahead.

\textbf{Unit ablation:} While SHAP provides a way to quantify the impact of hidden units on the prediction of a particular metadata concept, it does not imply causality. To identify causality we perform an ablation and measure the impact on the evaluation metrics. We remove units relevant to $\texttt{visible}\,_{t}$ and $R_{2}\theta_{000}$ prediction and measure the impact on the model's performance in terms of SPL, success, and episode length. We compare the ablation results to removing a random selection of units as a baseline. To remove a unit, we set the unit's activity as a constant that is equal to the mean of that unit's activity over the training episodes.

In Figure~\ref{fig6}, we observe that removing only 10 target units leads to a huge drop in SPL as compared to removing as many as 50 random units or units encoding reachability. As we remove more target units, the success also begins to drop. This suggests that target units are crucial and removing them first deteriorates the agents ability to identify targets thus leading to longer episodes and low SPL scores and beyond a certain point, the agent ability to be successful is also affected. Removing reachability units also leads to drop in SPL but the impact is not as drastic as in the case of target units. Interestingly removing reachability units lead to increase in success rate potentially due to an increase in exploration. Removing randomly selected units do not significantly impact any of the performance measures.

\begin{figure}
  \centering
  \includegraphics[width=1\linewidth]{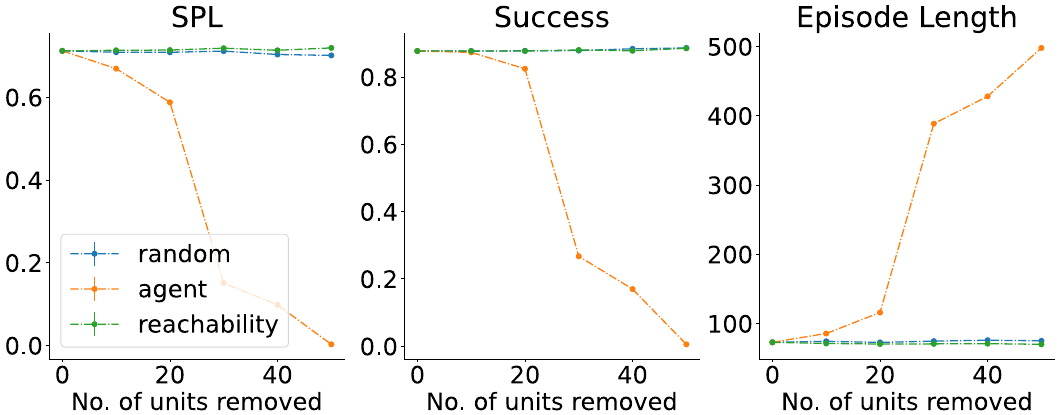}
    \vspace{-0.5cm}
  \caption{\textbf{Impact of removing units from \scpn.}}
    \vspace{-0.5cm}
  \label{fig8}
\end{figure}

\subsection{\pnav}
\vspace{-0.1cm}
Similar to \onav, we choose checkpoints after 50 million steps for our \pnav\ models. \rnpn\ (success = 0.925, SPL = 0.755) and \scpn\ (success = 0.878, SPL = 0.712) are highly successful at this task. We consider concepts derived from metadata that are related to target information ($R_{t}$,$\theta_{t}$) , reachability ($R_{r}\theta_{angle}$ where $r$ is the radius and angle is the orientation of the neighboring grid point with respect to the agent), agent's information ($R_{a}$,$\theta_{a}$) and visited history ($\texttt{visited}\,_{l}$,$\texttt{visited}\,_{lr}$,$\texttt{visited}\,_{lrh}$).

\textbf{Metadata prediction:} We train the GBTs to predict metadata from the GRU units. We first observe from Figure \ref{fig7}a (left) that reachability is predicted at all the angles well. Another interesting thing to note is that models that are not even trained on the \pnav\ task (\rnpnrnd\ and \scpnrnd) can predict reachability. This result is surprising as compared to \onav, where the only model that predicted reachability well was the one that performed well on the \onav\ task (\rnon). Further, \rnon\ only predicted the reachability in the view of the agent. Our intuition for the above result is that this could be due to additional information from GPS + compass sensor that provides the distance and orientation of the target. To tease apart the prediction due to visual sensor and GPS sensor we perform an ablation study where in one case we replace the output of the GPS sensor by random noise (visual-only; Figure \ref{fig7}a center)  and in the other we replace the image with all zeros (only GPS; Figure \ref{fig7}a right). 

In the visual-only case, we now observe a pattern similar to \onav\ where reachability in the field of view is more predictable than out of view. However, it is important to note that prediction of reachability does not improve with training \rnpn\ suggesting that ResNet with ImageNet weights is sufficient to predict reachability required to solve \pnav. \scpnrnd\ however does not seem to predict front reachability ($R_{2}\theta_{000}$) as effectively as \scpn\ suggesting that a random initialization is not sufficient to predict reachability required to solve \pnav. 

In the target-only case, we observe that the reachability of the backside of the agent is more predictable compared to the angles in the field of view. One possible explanation for this could be that when the distance between target and the agent changes in a given step that means the position at the back was reachable since the agent was there in the previous step. Therefore, using the change in GPS sensor values reachability at back can be predicted in some cases. 

The target distance and orientation is predictable when the GPS sensor is available for all the models (Figure \ref{fig7}b and Appendix \ref{A3}). This finding is expected as we provide this information as input, and when the GPS sensor is noise it can not be predicted (refer to Appendix \ref{A3}). Interestingly when the GPS sensor is available (Figure \ref{fig7}c and Appendix \ref{A3}), hidden units in trained \pnav\ models can predict the distance of the agent ($R_{a}$) from the initial spawn location. When using the SHAP method to find the relevant units for predicting $R_{a}$, we observe that top most relevant units have a constant value (refer to Appendix \ref{A4}) at almost every step in the episode and show very low variance in its output. On further inspection, we found that the 2 units in top-20 most relevant units for $R_{a}$ prediction were also relevant for target distance $R_{t}$ prediction. To predict $R_{a}$, GBT might be using a combination of a constant unit(s) and unit that encodes the target information.

\textbf{Unit ablation:} Similar to \onav\ we perform ablations by removing units and measuring the impact on the metrics. As shown in Figure~\ref{fig8} removing random and reachability units have almost no impact on the performance. Even after removing 50 units we observe similar performance on all three metrics. Removing the units that are relevant for predicting $R_{a}$ causes a significant drop in the performance and on dropping 50 units both SPL and success rate almost reach zero. The episode length also reaches the highest possible value (500) set in the task definition i.e. the episode ends if agent takes 500 steps. On further inspection, we found that in top-50 $R_{a}$ units, there are 6 units from the top-50 $R_{t}$ units. This is the key reason why \pnav\ performance dropped as the target distance information is lost. We further performed an ablation by removing only these 6 target units, which resulted in a drastic drop.

\vspace{-0.2cm}
\section{Conclusion}
\vspace{-0.1cm}
\label{sec:conclusion}
 We propose \framework to investigate if concepts about the agent, environment and task are encoded in the hidden representation of embodied agents. While we focus on visual navigation agents trained in AI2-THOR, the framework is generic and can be applied to agents trained on any task in any virtual environment with relevant metadata available. Our analysis shows the \onav agent encodes target orientation, reachability and visited locations history in order to avoid obstacles and visiting the same locations repeatedly. \pnav agents encode target orientation and its progress towards the target and show less reliance on visual information.


{\small
\bibliographystyle{ieee_fullname}
\bibliography{egbib}
}
\newpage
\setcounter{figure}{0}
\renewcommand{\thefigure}{A\arabic{figure}} 
\appendix
\section*{Appendix}
\begin{figure*}
  \centering
  \includegraphics[width=0.9\linewidth]{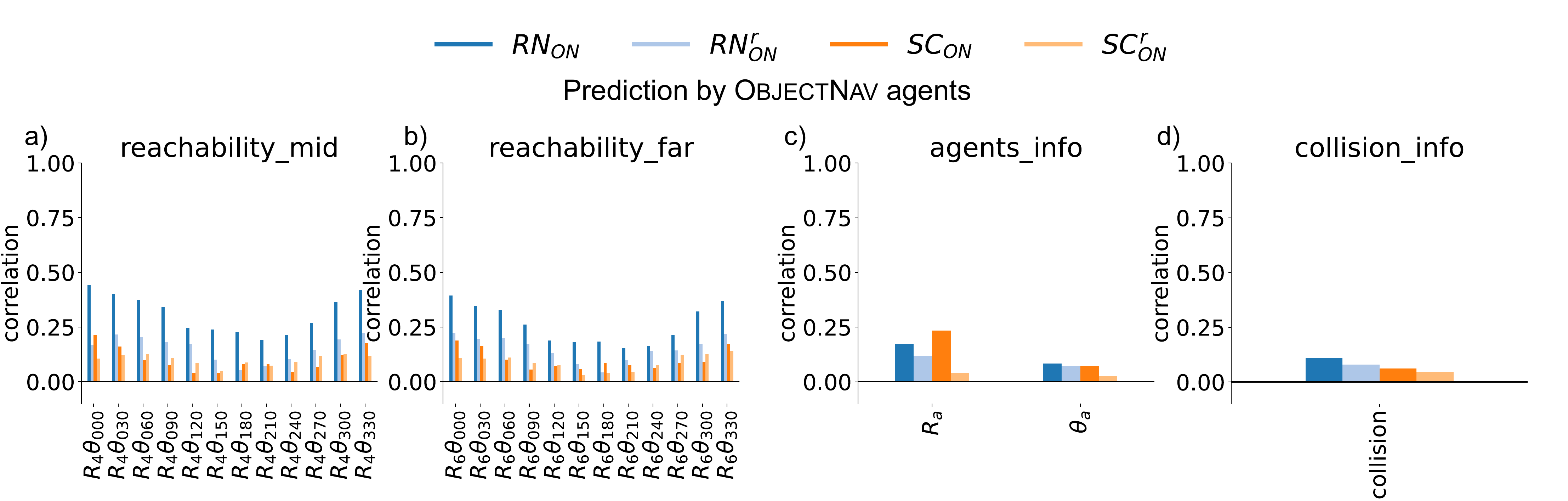}
  \caption{{\textbf{Metadata prediction by \onav GRU units:} a) Reachability mid (R = $4\times gridSize$) b) Reachability far (R = $6\times gridSize$)  c) Agent information d) Collision} }
  \label{sfig1}
\end{figure*}

\begin{figure*}
  \centering
  \includegraphics[width=1\linewidth]{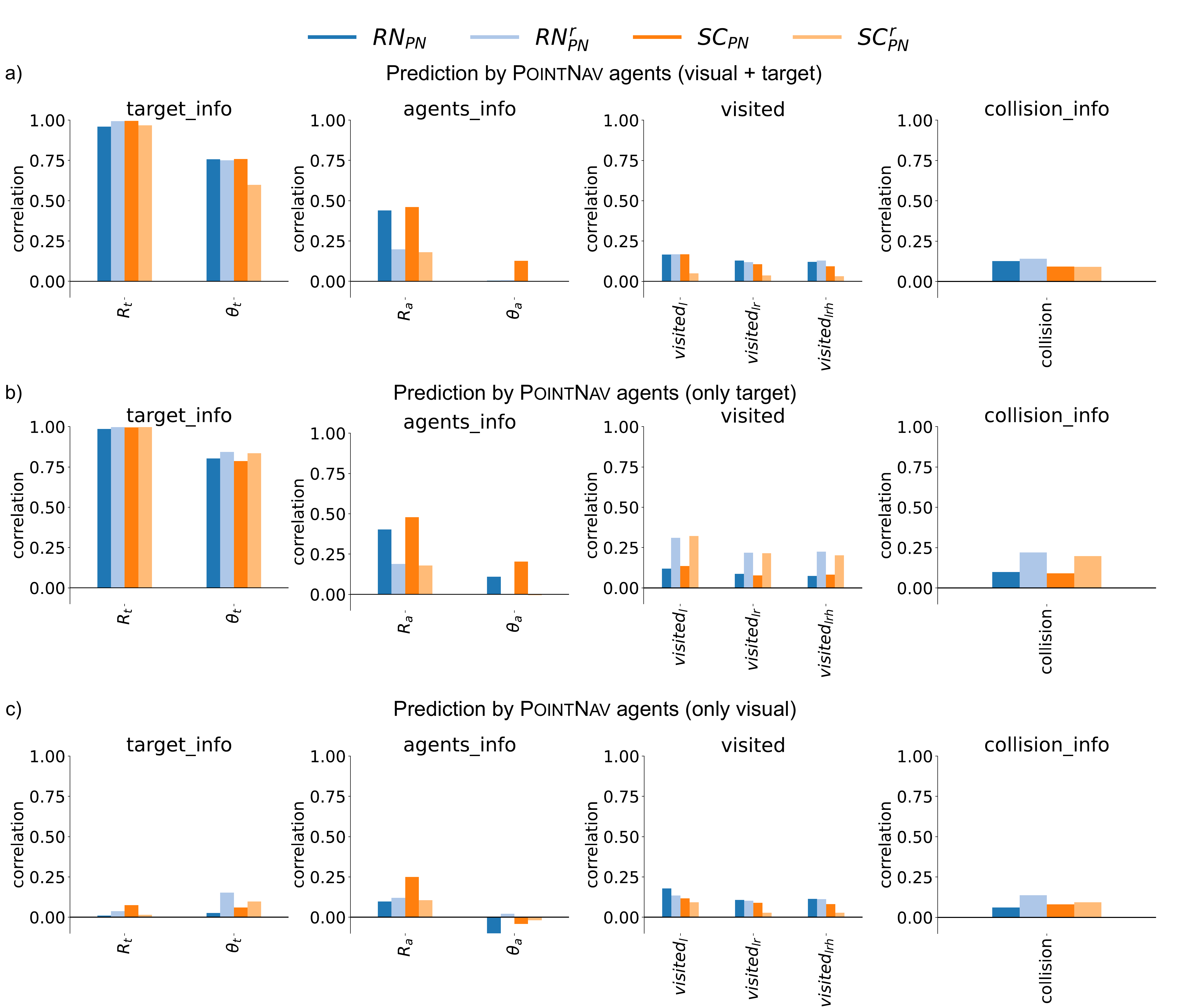}
  \caption{\textbf{Metadata prediction by \pnav GRU units:} using a) both visual and target sensor b) only target sensor c) only visual sensor  }
  \label{sfig2}
\end{figure*} 
\section{SHapley Additive exPlanations (SHAP)}
\label{A1}
In \framework, we apply SHAP to identify which GRU units were relevant in the prediction of a concept of interest. Let $f$ denotes the model (Gradient boosted tree) that is trained to predict the concept from hidden units. Let $S$ denote a subset of GRU units, then $f_x(S) \approx E[f(x) \mid x_S]$  is the estimated expectation of the model's output conditioned on the set S of GRU units. Then, relevance of a GRU unit $i$ (Shapley value: $\phi_i(f,x)$) in predicting the concept for a given example is given by

\begin{equation}
\phi_i(f,x) = \sum_{R \in \mathcal{R}} \frac{1}{M!} \left [ f_x(P^R_i \cup i) - f_x(P^R_i) \right ]
\label{eq:shapley}
\end{equation}

where $\mathcal{R}$ is the set of all unit orderings, $P^R_i$ is the set of all GRU units that come before unit $i$ in ordering $R$, and $M$ is the number of GRU units. In simple words, we calculate the change in outcome of model $f$ when the unit $i$ is added in the model with existing GRU units subset $P^R_i$. Then by averaging the change over all possible subsets, we get the Shapley value of unit $i$ for a given example thus providing the local importance for that example. To obtain the global importance of unit $i$, we aggregate Shapley values of unit $i$ over multiple examples from the validation trajectories. 

Although computing exact Shapley values in model agnostic setting  has exponential time complexity, Lundberg et al.\cite{lundberg2020local2global} came up with an elegant algorithm for GBTs that allows computing Shapley values in polynomial time. We refer the reader to \cite{lundberg2020local2global} for more details about the efficient SHAP algorithm for GBTs.  

\section{Concept prediction by \onav agent}
\label{A2}
We report the prediction results of concepts that were not reported in the main text in Figure \ref{sfig1}. We observe that reachability at radius = $4 \times gridSize$ (Figure \ref{sfig1}a) and radius = $6 \times gridSize$ (Figure \ref{sfig1}b) shows a pattern similar to radius = $2 \times gridSize$ as reported in the main text. Reachability of angles in front (around 0 degrees) is more predictable than angles in back (around 180 degrees) of the agent. The agent's position ($R_{a}$) with respect to its spawn location is only slightly more predictable than baselines while orientation ($\theta_{a}$) prediction is almost equal to baselines (Figure \ref{sfig1}c). The prediction of collision event is also not much better than baselines (Figure \ref{sfig1}d). 

\begin{figure*}
  \centering
  \includegraphics[width=1\linewidth]{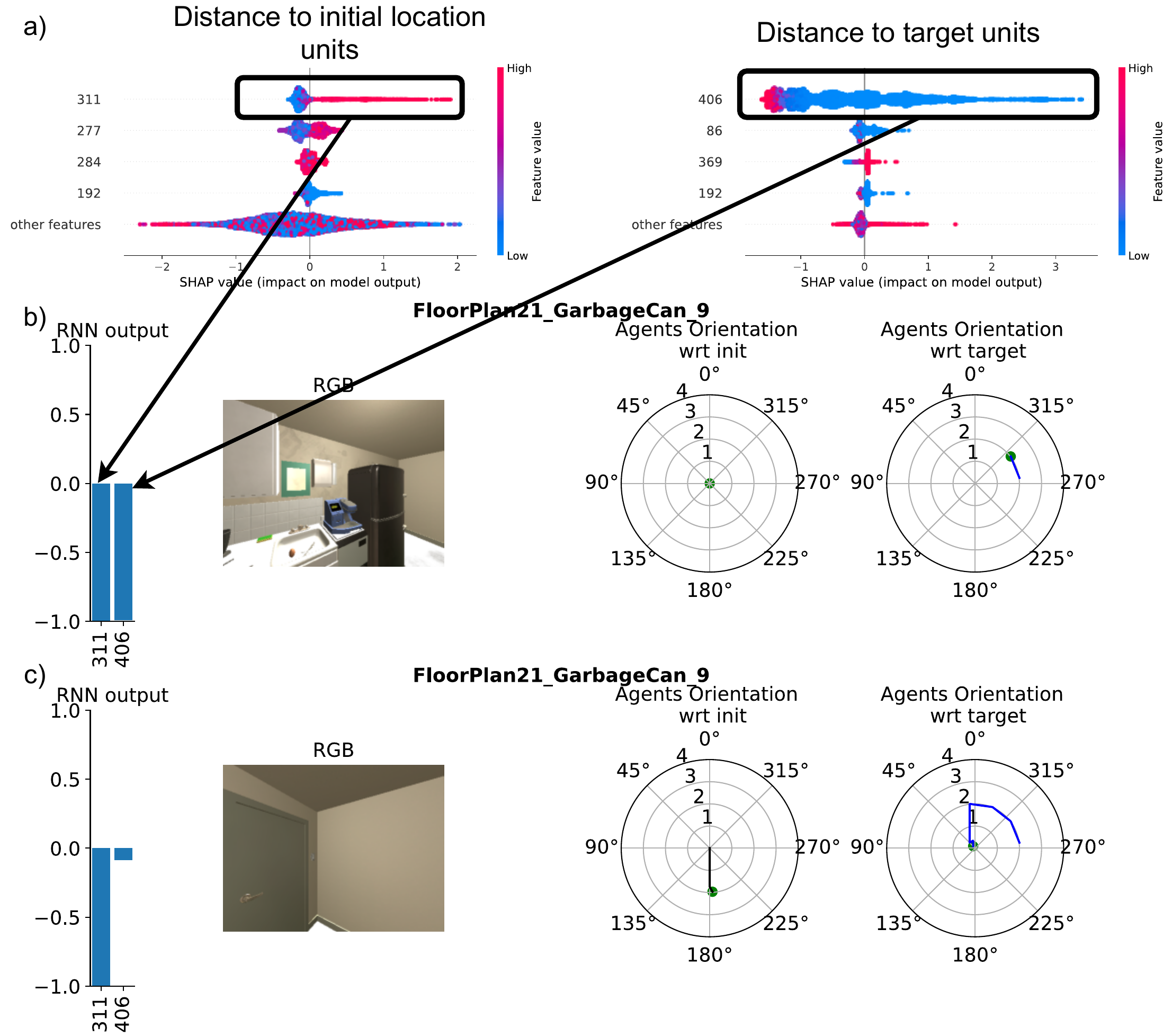}
  \caption{\textbf{Visualization of \pnav hidden units} a)   Top-4 most relevant hidden units for prediction of distance to agent's initial location ($R_{a}$) and target location ($R_{t}$) b) The bar plot on left shows response of unit 311 ($R_{a}$ unit) and unit 406 ($R_{t}$ unit). The image at the center is agent's current observation. The polar plots on right shows the distance (in meters) and orientation of the agent (in degrees) with respect to agent's initial location (third column) and with respect to target (fourth column). In this case, the agent is at around 2 meters away from the target and is oriented around 315 degrees. The response of both the units is negative. c) In this case, the agent is now closer to target and unit 406's ($R_{t}$ unit) response increases significantly suggesting that unit 406 is negatively correlated to $R_{t}$. Unit 311's response remains almost constant throughout the episode. We further found that unit 406 was in $11^{th}$ most relevant unit for predicting $R_{a}$ suggesting that a constant unit like unit 311 together with a $R_{t}$ unit (e.g. 406) is predicting $R_{a}$.}
  \label{sfig3}
\end{figure*}

\begin{figure*}
  \centering
  \includegraphics[width=1\linewidth]{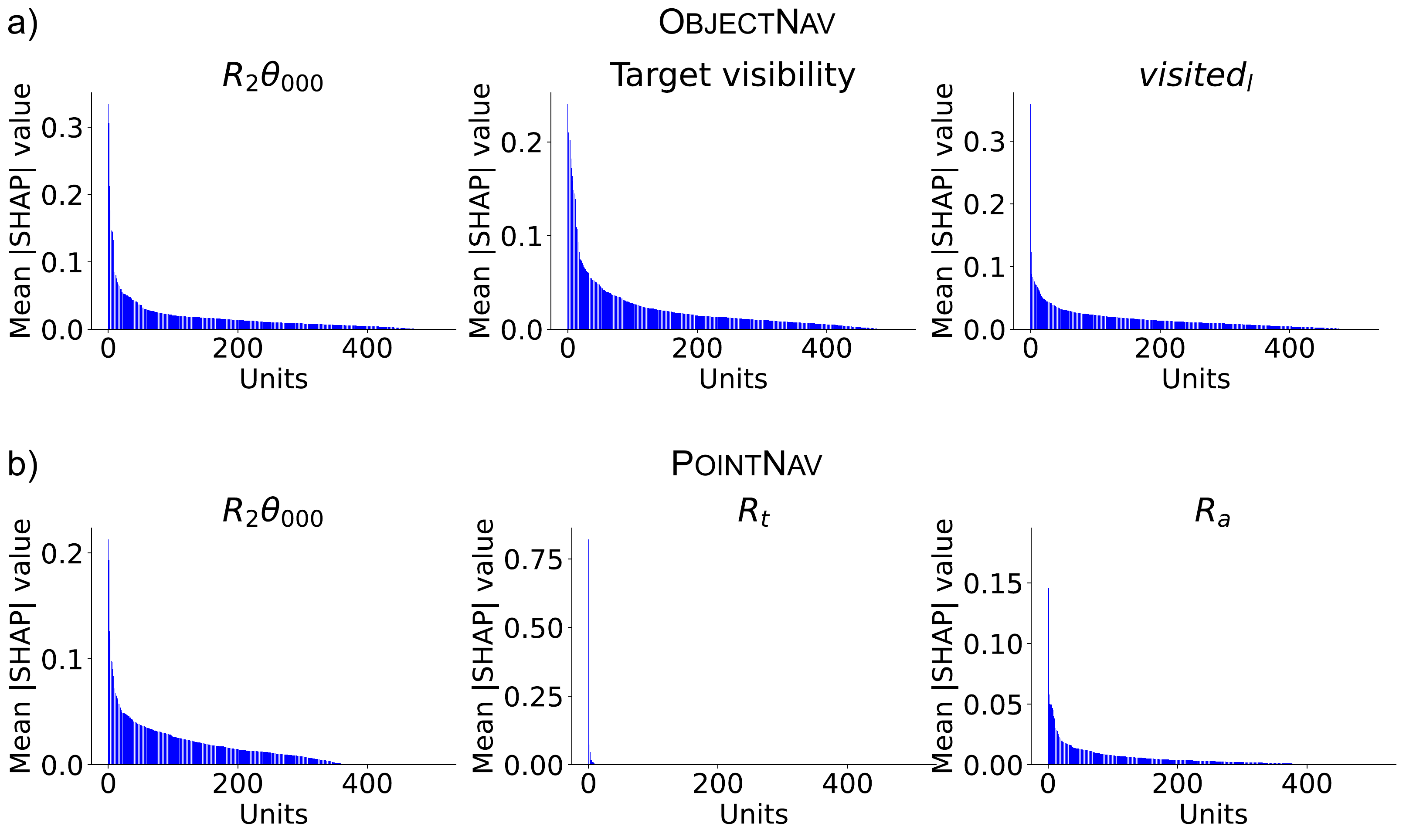}
  \caption{\textbf{Distribution of aggregate SHAP values:} for a) concepts learned by \onav agent and b) concepts learned by \pnav agent. The units are ordered in the decreasing order of the aggregate SHAP values.  }
  \label{sfig4}
\end{figure*}

\section{Concept prediction by \pnav agent}
\label{A3}
We report the prediction results of concepts that were not reported in the main text in Figure \ref{sfig2}. We observe that when GPS sensor (target distance and orientation) is available all models can predict target distance and orientation well. The agent's position with respect to its spawn location can be predicted by \pnav agents but not the baselines when GPS sensor is available. Further, when the GPS sensor is removed, the agent's position and orientation can not be predicted. Visited history and collision events are also not predictable. Surprisingly, visited history and collision events are slightly more predictable for baselines than \pnav trained models when only target information is used. A possible explanation for above observation could be that the position and orientation of a location  with respect to target will be same if a location is visited (or in case of a collision event) and therefore can be predicted using the GPS sensor. However, since the trained \pnav models do not predict visited history and collision events , this information might not be relevant to solving \pnav task.  

\section{Visualization of $R_{a}$ and $R_{t}$ units of \scpn}
\label{A4}
In Figure \ref{sfig3} a, we visualize the SHAP plots for top-4 units most relevant for predicting agent's position with respect to spawn location ($R_{a}$) and target's position with respect to agent's current location ($R_{t}$) as these were most predictable concepts from \scpn 's hidden state. From Figure \ref{sfig3} b and c we observe that top $R_{a}$ unit is constant throughout the episode while top $R_{t}$ unit's response increases as the agent moves closer to the target suggesting that $R_{t}$ unit's response is negatively correlated with target distance. As top $R_{a}$ unit's response was almost constant throughout the episode we further investigated its variance in all the validation episodes and observed that it was extremely low ($4.9 \times 10^{-9}$). This result was surprising as we expected it to be correlated with agent's progress towards the target. One possibility is that the change in hidden's unit response is extremely low and it is not possible to visualize its change with respect to other units. Another possibility is that this unit might not be encoding $R_{a}$ independently but combined with other units (e.g. $R_{t}$ units). The main focus of \framework was on identifying individual units relevant for predicting a concept and in current form it can not explain how multiple units together predict a concept. 

\section{SHAP value distribution across units}
\label{A5}
In the main text, we visualized the distribution of SHAP values across individual examples in the validation set for units most relevant for predicting a concept. In Figure \ref{sfig4}, we visualize the distribution of aggregate (average of absolute SHAP values across examples) SHAP values across units for the most predictable concepts by \onav (Figure \ref{sfig4}a) and \pnav (Figure \ref{sfig4}b) agents. We observe that generally for all concepts, aggregrate SHAP values show a sharp drop suggesting only a few units are relevant for predicting a concept. 

\begin{figure*}
  \centering
  \includegraphics[width=1\linewidth]{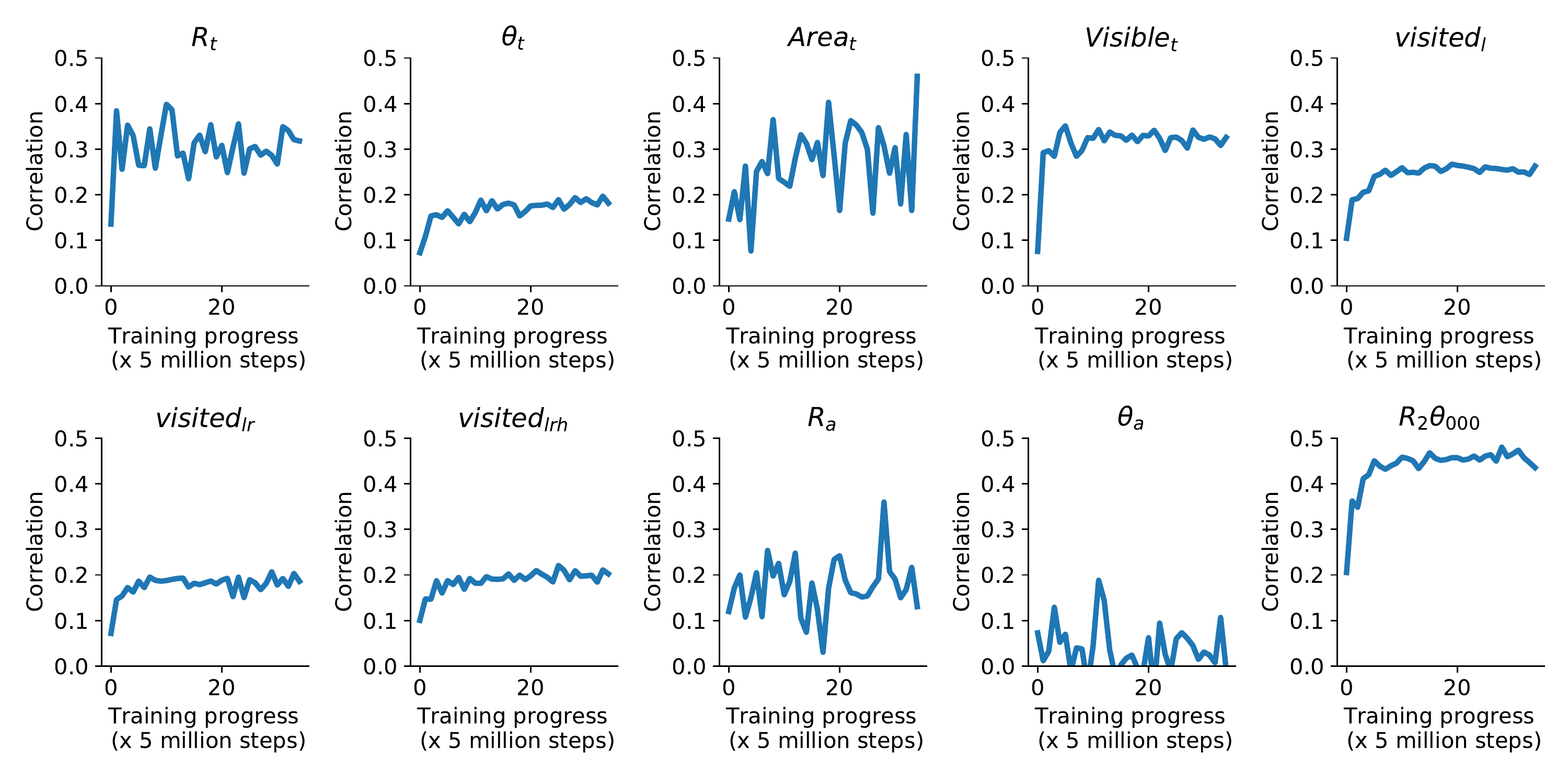}
  \caption{\textbf{Concept prediction vs. \rnon training progress}  }
  \label{sfig5}
\end{figure*}

\section{Concept prediction vs. \onav training progress}
\label{A6}
We evaluate which concepts investigated in this work were better predicted as the training progressed. In Figure \ref{sfig5}, we observe that prediction of target visibility, visited history and reachability improves significantly as compared to other concepts suggesting their importance for \onav task.  

\section{Irrelevant units ablation}
\label{A7}
We investigated removing units that were not relevant for predicting any learned concept in \onav(\rnon) and \pnav(\scpn). The table below shows the impact of removing 25\% and 50 \% of the irrelevant units. As we can observe removing 25\% units does not impact performance significantly in \onav. In \pnav, although the success is close even after dropping 25\% of the irrelevant units SPL drops significantly. Dropping more units (50\%) significantly drops performance in both \onav and \pnav tasks suggesting that we did not exhaustively investigate all possible concepts that were relevant for performing these tasks and the ablated units might be encoding those missing concepts.

\begin{center}
\footnotesize
\begin{tabular}{ ccccc } 
 \hline
 & \multicolumn{2}{c}{\textbf{ObjectNav}} & \multicolumn{2}{c}{\textbf{PointNav}}\\
 \hline
 Ablated units & SPL & Success & SPL & Success \\
 \hline
 0\% & 0.227 & 0.455 & 0.714 & 0.879 \\
 25\% & 0.225 & 0.445 & 0.659 & 0.865 \\ 
 50\% & 0.204 & 0.419 & 0.314 & 0.578 \\ 
 \hline
\end{tabular}
\end{center}

\section{Limitations and Future directions}
\label{A8}
This study has investigated several human interpretable concepts such as target visibility, reachability, etc. However, one can continue to broaden the set of concepts considered in such a study. We also restrict this study to navigation agents, but future studies should consider agents performing interactive tasks. Furthermore, we study RNN neurons individually. However, there is also a chance that multiple neurons together can encode some interesting property. We leave this study for future work.

The insights gained from our paper can also benefit future works: 1. Sparse representation of the target and ablation experiments suggest that irrelevant neurons can be assigned to another task leading to an efficient multitask agent or removed to reduce the size.  2. We found \pnav  agents rely less on RGB information. That could be the reason why they perform well in unseen rooms. \onav agents reliance on RGB information could be a weakness and it might help to design separate modules for target identification and navigation to be more robust to room changes. 3. We found that during training, early models (lower performance) do not predict reachability, target visibility, and visited history as well as saturated models (higher performance) suggesting their importance for \onav tasks.

\paragraph{Licenses for assets}
In this work we use three publicly available assets:
\begin{itemize}
    \item AI2Thor\footnote{https://github.com/allenai/ai2thor/blob/main/LICENSE}: Apache 2.0 License
    \item Allenact\footnote{https://allenact.org/LICENSE/}: MIT License
    \item shap\footnote{https://github.com/slundberg/shap/blob/master/LICENSE}: MIT License
    \item xgboost\footnote{https://github.com/dmlc/xgboost}: Apache-2.0 License
    
\end{itemize}
\end{document}